\documentclass{llncs}

\usepackage{times}
\usepackage[english]{babel} 
\usepackage[all]{foreign} 
\usepackage[inline]{enumitem} 
\usepackage{comment} 
\usepackage{url}
\usepackage{graphicx}
\usepackage{amsmath}
\allowdisplaybreaks 
\usepackage{amssymb}
\usepackage{subcaption} 
\usepackage{wrapfig} 
\usepackage{tikz} 
\usepackage{booktabs} 
\usepackage{tabularx} 
\usepackage{pifont} 

\usepackage{etoolbox}
\AtBeginEnvironment{tabular}{\tiny}

\DeclareMathOperator*{\argmax}{arg\,max}

\definecolor{forestgreen}{RGB}{35, 142, 35}
\definecolor{lightred}{RGB}{237, 41, 57}

\usepackage{todonotes,regexpatch}
\makeatletter
\xpatchcmd{\@todo}{\setkeys{todonotes}{#1}}{\setkeys{todonotes}{inline,#1}}{}{}
\makeatother

\newcommand{\matthew}[1]{\todo[backgroundcolor=lightred]{{Matthew: #1}}}

\newcommand{\real}{{\mathbb R}}
\newcommand{\vect}[1]{\mathbf{#1}}
\newcommand{\ball}[2]{\mathbb{B}(#1,#2)}

\newcommand{\size}[1]{|#1|}
\newcommand{\indicator}[1]{\mathbb{I}_{#1}}
\newcommand{\distance}[2]{|#1 - #2|}
\newcommand{\outputdistance}[2]{||#1 - #2||}

\newcommand{\losssymbol}{\mathcal{L}}
\newcommand{\lossfn}{\losssymbol}

\newcommand{\dataset}{\mathcal{X}}

\newcommand{\sample}[2]{S(#1, #2)}

\newcommand{\x}{\vect{x}} 		
\newcommand{\xt}{\hat{\x}} 		
\newcommand{\xs}{\x} 			

\newcommand{\y}{\vect{y}} 		

\newcommand{\mnistfunc}{f_\text{MNIST}}

\newcommand{\propertyp}{P}            

\newcommand{\SR}[2]{SR(#1, #2)} 
\newcommand{\LR}[2]{LR(#1, #2)} 
\newcommand{\CR}[1]{CR(#1)} 
\newcommand{\SCR}[2]{SCR(#1,#2)} 

\DeclareMathOperator{\CL}{CAcc} 
\DeclareMathOperator{\CSec}{CSec} 
\DeclareMathOperator{\CS}{CSat} 

\newcommand{\yes}{\textcolor{forestgreen}{\ding{51}}}
\newcommand{\no}{\textcolor{red}{\ding{55}}}
\newcommand{\good}[1]{\textcolor{forestgreen}{#1}}
\newcommand{\average}[1]{\textcolor{orange}{#1}}
\newcommand{\bad}[1]{\textcolor{red}{#1}}

\newcommand{\redacted}[1]{URL redacted.}

\begin{document}

\title{Neural Network Robustness as a Verification Property:
A Principled Case Study
}

\author{
	Marco Casadio\inst{1}\and
	Ekaterina Komendantskaya\inst{1}\and
	Matthew L. Daggitt\inst{1}\and
	Wen Kokke\inst{2}\and
	Guy Katz\inst{3}\and
	Guy Amir\inst{3}\and
	Idan Refaeli\inst{3} }
\institute{
	Heriot-Watt University, Edinburgh, UK\\
	\email{\{mc248,ek19,md2006\}@hw.ac.uk}\and
	University of Strathclyde, Glasgow, UK\\
	\email{wen.kokke@strath.ac.uk}\and
	The Hebrew University of Jerusalem, Jerusalem, Israel\\
	\email{\{guykatz, guyam, idan0610\}@cs.huji.ac.il}
	}

\maketitle

\begin{abstract}
Neural networks are very successful at detecting patterns in noisy
data, and have become the technology of choice in many fields. However,
their usefulness is hampered by their susceptibility
to \emph{adversarial attacks}. 
Recently, many methods for measuring and improving a network's robustness to
adversarial perturbations have been proposed, and this growing body of
research has given rise to numerous explicit or implicit notions of
robustness. Connections between these notions are often subtle, and a
systematic comparison between them is missing in the literature.  
%
%
In this paper we begin addressing this gap, by
setting up general principles for the empirical 
analysis and evaluation of a network's robustness as a mathematical
property --- during the network's training phase, its verification, and after
its deployment.  
We then apply these principles and conduct a case study that 
showcases the practical benefits of our general approach. 


\end{abstract}


\section{Introduction}
\label{sec:introduction}



Safety and security are critical for many complex systems that use
deep neural networks (DNNs).
Unfortunately, due to the opacity of DNNs, these properties are difficult to ensure.
Perhaps the most famous instance of this problem is guaranteeing the
robustness of DNN-based systems against \emph{adversarial attacks}~\cite{intriguing_properties_of_neural_networks_2014_szegedy,GoodfellowSS14}.
Intuitively, a neural network is \emph{$\epsilon$-ball robust} around
a particular input if, when you move no more than~$\epsilon$ away from
that input in the input space, the output does not change much; or,
alternatively, the classification decision that the network gives does
not change. 
%
Even highly accurate DNNs will often display only low 
robustness, and so 
measuring and improving the adversarial robustness of DNNs has
received significant attention by both the machine learning and
verification communities~\cite{SinghGPV19,HuangKWW17,KatzHIJLLSTWZDK19}.


\begin{wrapfigure}{R}{0.55\linewidth}
  	\vspace{-1em}
 \begin{tikzpicture}[scale=.45]
 
\draw[fill=gray,draw=gray] (-3.95,.45) rectangle (1.05,-.55);  
\draw[fill=white] (-4,.5) rectangle (1,-.5); 
\node (0,0) { };
 \draw (-1.5,0) node {{\footnotesize \it Neural Network} };

\draw[fill=gray,draw=gray] (6.95,.45) rectangle (9.30,-.55);  
\draw[fill=white] (6.5,.5) rectangle (9.25,-.5); 
\draw (7.8,0) node{{\footnotesize \it Verifier}};

\draw (3.8,0.3) node{\footnotesize{\textbf{Verification}}};
\draw (3.8,-0.5) node{\footnotesize{\textbf{Property}}};

\draw[latex-,shorten <=2pt,shorten >=2pt,dashed] (6.9,.4) .. controls (5,2) and (3,2) .. (1,.4); 
\draw (4,2.3) node[anchor=north,fill=white]{\emph{\footnotesize{\textcolor{red}{verify}}}};
 
\draw[latex-,shorten <=2pt,shorten >=2pt,dashed] (1,-.4) .. controls (3,-2) and (5,-2) .. (6.9,-.4); 
\draw (4,-2.3) node[anchor=south,fill=white]{\emph{\footnotesize{\textcolor{red}{retrain}}}}; 
\end{tikzpicture}\caption{\footnotesize{\emph{Continuous Verification Cycle}}}\label{fig:CV}
	\vspace{-1em}
\end{wrapfigure}
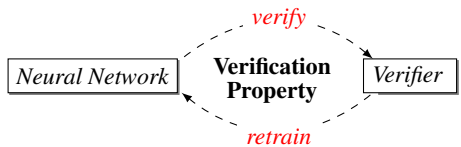

As a result, neural network verification
often follows a \emph{continuous verification cycle}~\cite{KKK20}, which involves retraining neural networks with a given  \emph{verification property} in mind, as Fig.~\ref{fig:CV} shows.
 More generally, such training can be regarded as a way to impose
a formal specification on a DNN; and so, apart from improving its
robustness, it may also contribute to the network's explainability,
and facilitate its verification. Due to the high level of interest in
adversarial robustness, numerous approaches have been proposed for
performing such retraining in recent years, each with its own specific
details. However it is quite unclear what are the benefits that
each approach offers, from a verification point of view.

The primary goal  of this case-study paper is to introduce a more
holistic methodology, which puts the
verification property in the centre of the development cycle,
and in turn permits a principled analysis of how this property 
influences both training and verification practices.
In particular, we
analyse the verification properties that implicitly or explicitly arise from the most prominent
families of training techniques:
\emph{data augmentation}~\cite{ShortenK19},
  \emph{adversarial training}~\cite{GoodfellowSS14,madry2018towards},
  \emph{Lipschitz robustness training}~\cite{anil2019sorting,pauli2021training},
 and  \emph{training with logical constraints}~\cite{XuZFLB18,FischerBDGZV19}.
We study the effect of each of these
properties on verifying the DNN in question.
%



In Section~\ref{sec:robustness-training}, we start with the forward direction of the continuous verification cycle, and show how
the above training methods give rise to logical properties of \emph{classification robustness} (CR),  \emph{strong classification robustness} (SCR), \emph{standard robustness} (SR) and \emph{Lipschitz robustness} (LR).
In Section~\ref{sec:robustness-definitions}, we trace the opposite direction of the cycle, i.e. show how and when the verifier failure in proving these properties can be mitigated. However Section~\ref{sec:experiments} first gives an auxiliary logical link for making this step. Given a robustness property as a logical formula, we can use it not just in verification, but also in attack or property accuracy measurements. 
We take property-driven attacks as a valuable tool in our study, both in training and in evaluation.
Section~\ref{sec:robustness-definitions} makes the underlying assumption that verification requires retraining: it shows that the verifier’s success ranges only 0 - 1.5\% for an accurate baseline network. 
We show how our logical understanding of robustness properties empowers us in property-driven training and in verification. We first give abstract arguments why certain properties are stronger than others or incomparable; and then we use training, attacks and the verifier Marabou to confirm them empirically.
%
%
Sections~\ref{sec:concl} and~\ref{sec:discussion} add other general considerations for setting up the continuous verification loop and conclude the paper.

\section{Existing Training Techniques and Definitions of Robustness}
\label{sec:robustness-training}

\textbf{Data augmentation} is a straightforward
method for improving robustness via training~\cite{ShortenK19}. It is
applicable to any transformation of the input (e.g. addition of noise,
translation, rotation, scaling) that leaves the output label
unchanged. To make the network robust against such a transformation,
one augments the dataset with instances sampled via the
transformation. 

More formally, given a neural network $N: \real^n \rightarrow
\real^m$, the goal of data augmentation is to ensure 
 \emph{classification robustness}, which is defined as follows. Given
 a training dataset input-output pair $(\xt,\y)$ and a distance metric
 $\distance{\cdot}{\cdot}$, for all inputs $\xs$ within the
 $\epsilon$-ball distance of $\xt$, we say that $N$ is
 \emph{classification-robust} if class $\y$ has the largest score in output $N (\xs)$. 
 \begin{definition}[Classification robustness]
  \begin{equation*}
    \label{eq:classification-robustness}
    \CR{\epsilon, \xt} \triangleq \forall \xs: \distance{\xs}{\xt} \leq \epsilon \Rightarrow \argmax N(\xs) = \y
  \end{equation*}

\end{definition}

%
In order to apply data augmentation, an engineer needs to specify: 
%
     \textbf{c1.}
    the value of $\epsilon$, i.e. the admissible range of perturbations;
%
  \textbf{c2.}
  the distance metric, which is determined according to the admissible
    geometric perturbations; and
%
      \textbf{c3.}
      the  sampling method used to produce the perturbed inputs (e.g.,
      random sampling, adversarial attacks, generative algorithm, prior knowledge of images).

      Classification robustness is straightforward, but does not
      account for the possibility of having ``uncertain'' images in
      the dataset, for which a small perturbation ideally should
      change the class. For datasets that contain a significant number
      of such images, attempting this kind of training could lead to a
      significant reduction in accuracy.

\textbf{Adversarial training} is a current state-of the-art method to robustify a neural network. Whereas standard training tries to minimise loss between the predicted value, $f(\xt)$, and the true value, $\y$, for each entry $(\xt, \y)$ in the training dataset, adversarial training minimises the loss with respect to the worst-case perturbation of each sample in the training dataset. It therefore replaces the standard training objective $\lossfn(\xt, \y)$ with:
$\max_{\forall \xs : \distance{\xs}{\xt} \leq \epsilon} \lossfn(\xs, \y)$.
Algorithmic solutions to the maximisation problem that find the worst-case perturbation has been the subject of several papers. The earliest suggestion was the Fast Gradient Sign Method (FGSM) algorithm introduced by \cite{GoodfellowSS14}: 
\begin{equation*}
\text{FGSM}(\xt) = \xt + \epsilon \cdot \text{sign}(\nabla_\x \lossfn(\x, \y))
\end{equation*}
However, modern adversarial training methods usual rely on some variant of the Projected Gradient Descent (PGD) algorithm \cite{madry2019deep}  which iterates FGSM: 
\begin{equation*}
	\text{PGD}_0(\xt) = \xt; \quad \text{PGD}_{t+1}(\xt) = \text{PGD}_{t}(\text{FGSM}(\xt))
\end{equation*}

It has been empirically observed that neural networks trained using this family of methods exhibit greater robustness at the expense of an increased generalisation error~\cite{tsipras2018robustness,madry2018towards,zhang2019theoretically}, which is frequently referred to as the \emph{accuracy-robustness trade-off} for neural networks (although this effect has been observed to disappear as the size of the training dataset grows~\cite{raghunathan2019adversarial}).

In logical terms what is this procedure trying to train for?
Let us assume that there's some maximum distance, $\delta$, that it is acceptable for the output to be perturbed given the size of perturbations in the input. This leads us to the following definition, where $\outputdistance{\cdot}{\cdot}$ is a suitable distance function over the output space:
\begin{definition}[Standard robustness]
  \begin{equation*}
    \label{eq:standard-robustness}
    \SR{\epsilon}{\delta, \xt} \triangleq \forall \xs: \distance{\xs}{\xt} \leq \epsilon \Rightarrow \outputdistance{f(\xs)}{f(\xt)} \leq \delta
  \end{equation*}
\end{definition}
We note that, just as with data augmentation, choices \textbf{c1} -- \textbf{c3} are still there to be made, although the sampling methods are usually given by special-purpose FGSM/PGD heuristics based on computing the loss function gradients.

\textbf{Training for Lipschitz robustness.}
More recently, a third competing definition of robustness has been proposed: Lipschitz robustness~\cite{balan2018lipschitz}. Inspired by the well-established concept of Lipschitz continuity, Lipschitz robustness asserts that the distance between the original output and the perturbed output is at most a constant $L$ times the change in the distance between the inputs.
\begin{definition}[Lipschitz robustness]
  \begin{equation*}
    \label{eq:lipschitz-robustness}
    \LR{\epsilon}{L, \xt} \triangleq \forall \xs: \distance{\xs}{\xt} \leq \epsilon \Rightarrow \outputdistance{f(\xs)}{f(\xt)} \leq L \distance{\xs}{\xt}
  \end{equation*}
\end{definition}
As will be discussed in Section~\ref{sec:robustness-definitions}, this is a stronger requirement than standard robustness. Techniques for training for Lipschitz robustness include formulating it as a semi-definite programming optimisation problem~\cite{pauli2021training} or including a projection step that restricts the weight matrices to those with suitable Lipschitz constants~\cite{gouk2021regularisation}. 

\textbf{Training with logical constraints.}
%
Logically, this discussion leads one to ask whether a more general approach to constraint formulation may exist, and
several attempts in the literature addressed this research question~\cite{XuZFLB18,FischerBDGZV19}, by proposing methods that can translate a first-order
logical formula~$C$ into a \emph{constraint loss function} $\lossfn_C$. The loss function penalises the network when outputs
do not satisfy a given Boolean constraint,
and universal quantification is handled by a choice of sampling method. Our standard loss function $\lossfn$ is substituted with:
	\vspace{-0.5em}
\begin{equation}
\lossfn^*(\xt, \y) = \alpha \lossfn (\xt, \y) + \beta \lossfn_C(\xt, \y)
\label{eqn:2}
\end{equation}
	\vspace{-0.1em}
where weights $\alpha$ and $\beta$ control the balance between the standard and constraint loss.

This method looks deceivingly as a generalisation of previous approaches.
However, even given suitable choices for \textbf{c1} -- \textbf{c3},
classification robustness cannot be modelled via a constraint loss in
the DL2~\cite{FischerBDGZV19} framework, as $argmax$ is not differentiable. Instead,
\cite{FischerBDGZV19} defines an alternative constraint, which we call \emph{strong classification robustness}:
\begin{definition}[Strong classification robustness]
  \begin{equation*}
    \label{eq:approximate-classification-robustness}
    \SCR{\epsilon}{\eta, \xt} \triangleq \forall \xs: \distance{\xs}{\xt} \leq \epsilon \Rightarrow f(\xs) \geq \eta
  \end{equation*}
\end{definition}
which looks only at the prediction of the true class and checks whether it is greater than some value~$\eta$ (chosen to be 0.52 in their work).


We note that sometimes, the constraints (and therefore the derived loss functions) refer to the true label $\y$ rather than the current output of the network $f(\xt)$, e.g.
 $ \forall \xs: \distance{\xs}{\xt} \leq \epsilon \Rightarrow \distance{f(\xs)}{\y} \leq \delta$.
This leads to scenarios where a network that \emph{is} robust around~$\xt$ but gives the wrong prediction, being penalised by $\lossfn_C$ which on paper is designed to maximise robustness. Essentially $\lossfn_C$ is trying to maximise both accuracy and constraint adherence concurrently. Instead, we argue that to preserve the intended semantics of $\alpha$ and $\beta$ it is important to instead compare against the current output of the network. 
Of course, this does not work for SCR because deriving the most
popular class from the output $f(\xt)$ requires the $\argmax$ operator
--- the very function that SCR seeks to avoid using. This is another argument why (S)CR should be avoided if possible.

\section{Robustness in Evaluation, Attack and Verification}\label{sec:experiments}


Given a particular definition of robustness, a natural question is how to quantify how close a given network is to satisfying it.
%
We argue that there are three different measures that one should be interested in:
 1. Does the constraint hold? This is a binary measure, and the answer is either true or false. 
2.  If the constraint does not hold, how easy is it for an attacker to find a violation?
3.  If the constraint does not hold, how often does the average user encounter a violation?
Based on these measures, we define three concrete metrics:
\emph{constraint satisfaction}, \emph{constraint security}, and \emph{constraint accuracy}.\footnote{Our naming scheme differs from \cite{FischerBDGZV19} who use the term \emph{constraint accuracy} 
  to refer to what we term \emph{constraint security}.
In our opinion, the term \emph{constraint accuracy}
is less appropriate here than
the name \emph{constraint security} given the use of an adversarial attack.}

Let $\dataset$ be the training dataset, 
$\ball{\xt}{\epsilon} \triangleq \{\x \in \real^n \mid \distance{\x}{\xt} \leq \epsilon\}$
be the $\epsilon$-ball around $\xt$ and~$\propertyp$ be the right-hand side of the implication in each of the definitions of robustness. 
Let $\indicator{\phi}$ be the standard indicator function which is 1 if constraint $\phi(\x)$ holds and~0 otherwise. The \emph{constraint satisfaction} metric measures the proportion of the (finite) training dataset for which the constraint holds.
\begin{definition}[Constraint satisfaction]
  	\vspace{-0.2em}
  \begin{equation*}
    \CS(\dataset) = \frac{1}{\size{\dataset}} \sum_{\xt \in \dataset} \indicator{\forall \xs \in \ball{\xt}{\epsilon}: \propertyp(\xs)}
  \end{equation*}
  
\end{definition}
\vspace{-0.4em}

\noindent In contrast, \emph{constraint security} measures the proportion of inputs in the dataset such that an attack $A$ is unable to find an adversarial example for constraint $P$. In our experiments we use the PGD attack for $A$, although in general any strong attack can be used.
\vspace{-0.2em}
\begin{definition}[Constraint security]
  \vspace{-0.2em}
  \begin{equation*}
    \CSec(\dataset) = \frac{1}{\size{\dataset}} \sum_{\xt \in \dataset} \indicator{\propertyp}(\text{A}(\xt))
  \end{equation*}
\end{definition}
\vspace{-0.4em}

Finally, \emph{constraint accuracy} estimates the probability of a random user coming across a counter-example to the constraint, usually referred as \emph{1 - success rate} in the robustness literature. Let $\sample{\xt}{n}$ be a set of $n$ elements randomly uniformly sampled from $\ball{\xt}{\epsilon}$. Then constraint accuracy is defined as:
\begin{definition}[Constraint accuracy]
  \begin{equation*}
    \CL(\dataset) = \frac{1}{\size{\dataset}} \sum_{\xt \in \dataset} \left(\frac{1}{n} \sum_{\xs \in \sample{\xt}{n}} \indicator{\propertyp}(\xs)\right)
  \end{equation*}
\end{definition}
Note that there is no relationship between constraint accuracy and constraint security: an attacker may succeed in finding an adversarial example where random sampling fails and vice-versa. Also note the role of sampling in this discussion and compare it to the discussion of the choice \textbf{c3} in Section~\ref{sec:robustness-training}.
Firstly, sampling procedures affect both training and evaluation of networks. But at the same time, their choice is orthogonal to choosing the verification constraint for which we optimise or evaluate.
For example, we measure constraint security with respect to the PGD attack, and this determines the way we sample; but having made that choice still leaves us to decide which constraint, SCR, SR, LR, or other we will be measuring as we sample. Constraint satisfaction is different from constraint security and accuracy, in that it must evaluate constraints over infinite domains rather than merely sampling from them.

\textbf{Choosing an evaluation metric.}
It is important to note that for all three evaluation metrics, one still has to make a choice for constraint $\propertyp$, namely SR, SCR or LR, as defined in Section~\ref{sec:robustness-training}. As constraint security always uses PGD to find input perturbations, the choice of SR, SCR and LR effectively amounts to us making a judgement of what an adversarial perturbation consists of: is it a class change as defined by SCR, or is it a violation of the more nuanced metrics defined by SR and LR? Therefore we will evaluate constraint security on the \emph{SR/SCR/LR constraints} using a \emph{PGD attack}. 

For large search spaces in $n$ dimensions, random sampling deployed in constraint accuracy fails to find the trickier adversarial examples, and usually has deceivingly high performance: we found $100\%$ and $>98\%$ constraint accuracy for SR and SCR, respectively. We will therefore not discuss these experiments in detail. 



\section{Relative Comparison of Definitions of Robustness}
\label{sec:robustness-definitions}

 \begin{wrapfigure}{R}{0.6\linewidth}
	\centering
	\begin{subfigure}{1\linewidth}
		\includegraphics[width=1\linewidth]{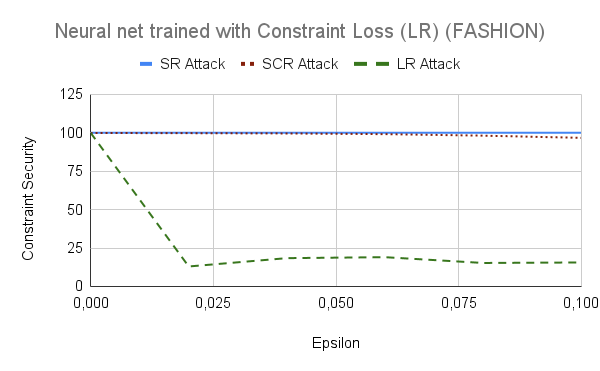}
	\end{subfigure}
	\hspace{0.03\textwidth}
	\begin{subfigure}{1\linewidth}
		\includegraphics[width=1\linewidth]{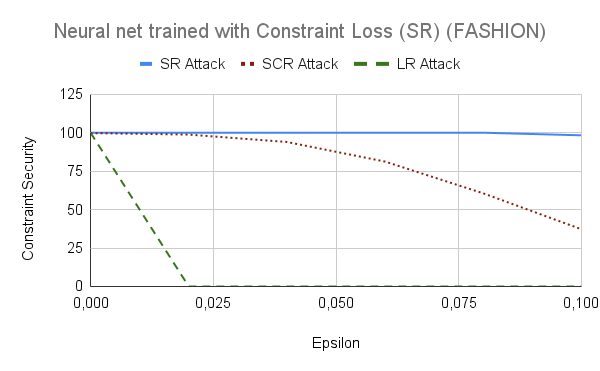}
	\end{subfigure}
	\vspace{-1em}
	\caption{\emph{\footnotesize{Experiments that show how the two networks trained with LR and SR constraints perform when evaluated against different definitions of robustness underlying the attack; $\epsilon$ measures the attack strength.}}}
	\label{fig:SL-LR}
	\vspace{-1.4em}
      \end{wrapfigure}

We now compare the strength of the given definitions of robustness  using the introduced metrics. 
For empirical evaluation, we train networks on \emph{FASHION MNIST} (or just \emph{FASHION})~\cite{fashionmnist} and a modified version of the \emph{GTSRB}~\cite{gtsrb} datasets consisting, respectively,  by 28x28 and 48x48 images belonging to 10 classes. 
The networks consist of two fully connected layers: the first one having 100 neurons and ReLU as activation function, and the last one having 10 neurons on which we apply a clamp function $[-100, 100]$, because the traditional softmax function is not compatible with constraint verification tools such as Marabou.
Taking four different robustness properties for which we optimise while training (Baseline, LR, SR, SCR), gives us 8 different networks to train, evaluate and attack. Generally, all trends we observed for the two data sets were the same, and we put matching graphs in~\cite{CDKK21} whenever we report a result for one of the data sets.  
Marabou~\cite{marabou} was used for evaluating constraint satisfaction. 

\subsection{Standard and Lipschitz Robustness}

Lipschitz robustness is a strictly stronger constraint than standard robustness, in the sense that when a network satisfies $\LR{\epsilon}{L}$ then it also satisfies $\SR{\epsilon}{\epsilon L}$. However, the converse does not hold, as standard robustness does not relate the distances between the inputs and the outputs. Consequently, there are $\SR{\epsilon}{\delta}$ robust models that are not $\LR{\epsilon}{L}$ robust for any $L$, as for any fixed $L$ one can always make the distance $\distance{\x}{\xt}$ arbitrarily small in order to violate the Lipschitz inequality.

\textbf{Empirical significance of the conclusions for constraint security.}
Fig.~\ref{fig:SL-LR} shows an empirical evaluation of this general result.
If we train two neural networks, one with the SR, and the other with the LR constraint, then the latter always has higher constraint security against both SR and LR attacks than the former. 
It also confirms that generally, stronger constraints are harder to obtain: whether a network is trained with SR or LR constraints, it is less robust against an LR attack than against any other attack.

\begin{table}[t]
\centering
\caption{\emph{\footnotesize{Constraint satisfaction results for the Classification, Standard and Lipschitz constraints. These values are calculated over the test set and represented as \%.}}}
\begin{tabularx}{\linewidth}{c|cccc|cccc}
\toprule
& \multicolumn{4}{c|}{FASHION net trained with:} & \multicolumn{4}{c}{GTSRB net trained with:} \\
& \ \ \ \ Baseline\ \ \ & \ \ \ SCR\ \ \ & \ \ \ SR\ \ \ & \ \ \ LR\ \ \ & \ \ \ \ Baseline\ \ \ & \ \ \ SCR\ \ \ & \ \ \ SR\ \ \ & \ \ \ LR\ \ \ \\ \midrule
\ CR satisfaction\ \ \ & \ \ \ 1.5\ \ \ & \ \ \ 2.0\ \ \ & \ \ \ 2.0\ \ \ & \ \ \ 34.0\ \ \ & \ \ \ 0.5\ \ \ & \ \ \ 1.0\ \ \ & \ \ \ 3.0\ \ \ & \ \ \ 4.5\ \ \ \\ \midrule
\ SR satisfaction\ \ \ & \ \ \ 0.5\ \ \ & \ \ \ 1.0\ \ \ & \ \ \ 65.8\ \ \ & \ \ \ 100.0\ \ \ \ & \ \ \ 0.0\ \ \ & \ \ \ 0.0\ \ \ & \ \ \ 24.0\ \ \ & \ \ \ 97.0\ \ \ \\ \midrule
\ LR satisfaction\ \ \ & \ \ \ 0.0\ \ \ & \ \ \ 0.0\ \ \ & \ \ \ 0.0\ \ \ & \ \ \ 0.0\ \ \ & \ \ \ 0.0\ \ \ & \ \ \ 0.0\ \ \ & \ \ \ 0.0\ \ \ & \ \ \ 0.0\ \ \ \\ \bottomrule
\end{tabularx}
\label{tab:marabou_results}
\vspace{-1em}
\end{table}

\textbf{Empirical significance of the conclusions for constraint satisfaction.}  
Table~\ref{tab:marabou_results} shows that LR is very difficult to guarantee as a verification property, indeed none of our networks satisfied this constraint for any image in the data set. At the same time, networks trained with LR satisfy the weaker property SR, for 100\% and 97\% of images -- a huge improvement on the negligible percentage of robust images for the baseline network! Therefore, knowing a verification property or mode of attack, one can tailor the training accordingly, and training with stronger constraint gives better results.

\begin{wrapfigure}{R}{1.5\linewidth}
	\centering
	\begin{subfigure}{1\linewidth}
		\includegraphics[width=1\linewidth]{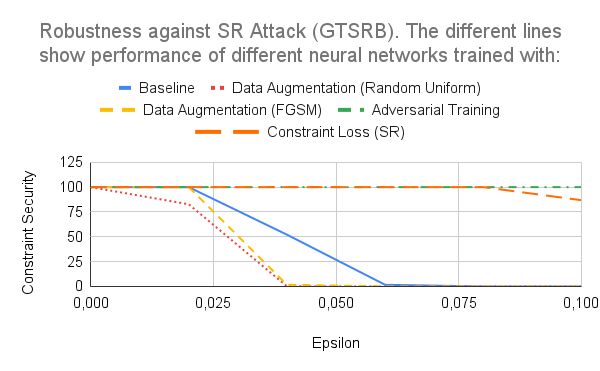}
	
	\end{subfigure}
	\hspace{0.03\textwidth}
	\begin{subfigure}{1\linewidth}
		\includegraphics[width=1\linewidth]{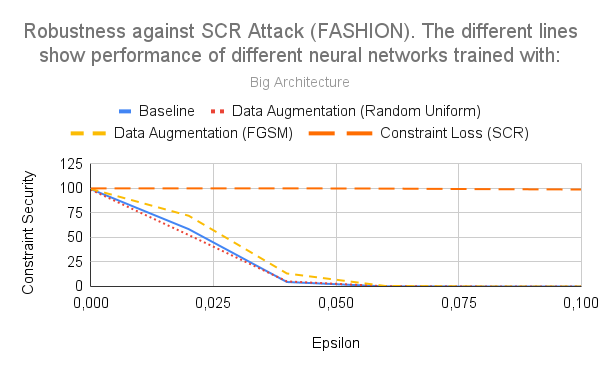}
              \end{subfigure}
%
	\vspace{-1em}
	\caption{\emph{\footnotesize{Experiments that show how adversarial training, training with data augmentation, and training with constraint loss affect standard and classification robustness of networks;  $\epsilon$ measures the attack strength. 
            }}}
	\label{fig:modes}
	\vspace{-1.4em}
\end{wrapfigure}

\subsection{(Strong) Classification Robustness}


Strong classification robustness is designed to over-approximate classification robustness whilst providing a logical loss function with a meaningful gradient. We work under the assumption that the last layer of the classification network is a softmax layer, and therefore the output forms a probability distribution. When $\eta > 0.5$ then any network that satisfies $\SCR{\epsilon}{\eta}$ also satisfies $\CR{\epsilon}$. For $\eta \leq 0.5$ this relationship breaks down as the true class may be assigned a probability greater than~$\eta$ but may still not be the class with the highest probability. We therefore recommended that one only uses value of $\eta > 0.5$ when using strong classification robustness (for example $\eta = 0.52$ in \cite{FischerBDGZV19}).

      \textbf{Empirical significance of the conclusions for constraint security.}
      Because the CR constraint cannot be used within a loss function, we use data augmentation when training to emulate its effect. 
      First, we confirm our assumptions about the relative inefficiency of using data augmentation compared to adversarial training or training with constraints, see Fig.~\ref{fig:modes}.
      Surprisingly, neural networks trained with data augmentation give worse results than even the baseline network.

As previously discussed, random uniform sampling struggles to find adversarial inputs in large searching spaces.
It is logical to expect that using random uniform sampling when training will be less successful than training with sampling that uses FGSM or PGD as heuristics. Indeed, Fig.~\ref{fig:modes} shows this effect for data augmentation. 


One may ask whether the trends just described would be replicated for more complex architectures of neural networks. In particular, data augmentation is known to require larger networks. By replicating the results  with a large, 18-layer convolutional network
from~\cite{FischerBDGZV19} (second graph of Fig.~\ref{fig:modes}), we confirm that larger networks handle data augmentation better, and that data augmentation affords improved robustness compared to the baseline.
Nevertheless, data augmentation still lags behind all other modes of constraint-driven training, and thus this major trend remains stable across network architectures.
The same figure also illustrates our point about the relative strength of SCR compared to CR: a network trained with data augmentation (equivalent to CR) is more prone to SCR attacks than a network trained with the SCR constraint. 

      \textbf{Empirical significance of the conclusions for constraint satisfaction.} Although Table~\ref{tab:marabou_results} confirms that training with a stronger property (SCR) does improve the constraint satisfaction of a weaker property (CR), the effect is an order of magnitude smaller than what we observed for LR and SR. Indeed, the table suggests that training with the LR constraint gives better results for CR constraint satisfaction. This does not contradict, but does not follow from our theoretical analysis. 

\subsection{Standard vs Classification Robustness}

Given that LR is stronger than SR and SCR is stronger than CR, the obvious question is whether there is a relationship between these two groups. In short, the answer to this question is no. In particular, although the two sets of definitions agree on whether a network is robust around images with high-confidence, they disagree over whether a network is robust around images with low confidence. We illustrate this with an example, comparing SR against CR. We note that a similar analysis holds for any pairing from the two groups.

\begin{wrapfigure}{R}{0.33\linewidth}
	\centering
	\begin{subfigure}{.4\linewidth}
		\includegraphics[width=1\linewidth]{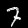}
		\caption{\tiny{$P(7) = 85\%$}}
		\label{fig:mnist-high-confidence}
	\end{subfigure}
	\hspace{0.03\textwidth}
	\begin{subfigure}{.4\linewidth}
		\includegraphics[width=1\linewidth]{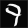}
		\caption{\tiny{$P(7) = 51\%$}}
		\label{fig:mnist-low-confidence}
	\end{subfigure}
	\caption{\emph{\footnotesize{Images from the MNIST set}}}
	\label{fig:mnist-high-low-confidence-examples}
	\vspace{-1em}
      \end{wrapfigure}

The key insight is that standard robustness bounds the drop in confidence that a neural network can exhibit after a perturbation, whereas classification robustness does not. 
Fig.~\ref{fig:mnist-high-low-confidence-examples} shows two hypothetical images from the MNIST dataset. Our network predicts that Fig.~\ref{fig:mnist-high-confidence} has an 85\% chance of being a 7.
Now consider adding a small perturbation to the image and consider two different scenarios.
In the first scenario the output of the network for class 7 decreases from 85\% to 83\% and therefore the classification stays the same.
In the second scenario the output of the network for class 7 decreases from 85\% to 45\%, and results in the classification changing from 7 to 9.
When considering the two definitions, a small change in the output leads to no change in the classification and a large change in the output leads to a change in classification and so robustness and classification robustness both agree with each other.

However, now consider Fig.~\ref{fig:mnist-low-confidence} with relatively high uncertainty. In this case the network is (correctly) less sure about the image, only narrowly deciding that it's a 7. Again consider adding a small perturbation. In the first scenario the prediction of the network changes dramatically with the probability of it being a 7 increasing from 51\% to 91\% but leaves the classification unchanged as~7. In the second scenario the output of the network only changes very slightly, decreasing from 51\% to 49\% flipping the classification from 7 to 9. Now, the definitions of SR and CR disagree. In the first case, adding a small amount of noise has erroneously massively increased the network's confidence and therefore the SR definition correctly identifies that this is a problem. In contrast CR has no problem with this massive increase in confidence as the chosen output class remains unchanged.
Thus, SR and CR agree on low-uncertainty examples, but CR breaks down and gives what we argue are both false positives and false negatives when considering examples with high-uncertainty.

 \begin{wrapfigure}{R}{0.65\linewidth}
	\centering
	\begin{subfigure}{1\linewidth}
		\includegraphics[width=1\linewidth]{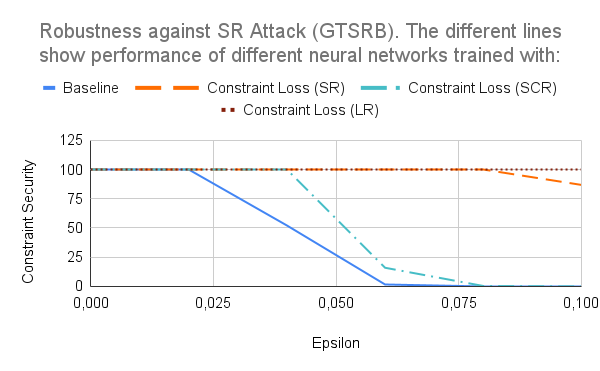}
	\end{subfigure}
	\vspace{-1em}
	\caption{\emph{\footnotesize{Experiments that show how different choices of a constraint loss affect standard robustness of neural networks.
            }}}
	\label{fig:CL}
	\vspace{-1.4em}
\end{wrapfigure}

\textbf{Empirical significance of the conclusions for constraint security.}
Our empirical study confirms these general conclusions.  Fig.~\ref{fig:SL-LR} shows that depending on the properties of the dataset, SR may not guarantee SCR.
The results in Fig.~\ref{fig:CL} tell us that using the SCR constraint for training does not help to increase defences against SR attacks.
A similar picture, but in reverse, can be seen when we optimize for SR but attack with SCR. 
Table~\ref{tab:marabou_results} confirms these trends for constraint satisfaction.

\section{Other Properties of Robustness Definitions}\label{sec:concl}

\newcommand{\wrapp}[2]{\begin{minipage}[t]{#1\columnwidth}%
    \centering #2
\end{minipage}}

\begin{table*}[t]
  \caption{\emph{\footnotesize{A comparison of the different types of robustness studied in this paper. Top half: general properties. Bottom half: relation to existing machine-learning literature}}}
  \begin{tabularx}{\linewidth}{lcccc}
    \toprule
    \small{Definition}
    & \wrapp{0.15}{\small{Standard robustness}}
    & \wrapp{0.15}{\small{Lipschitz robustness}}
    & \wrapp{0.2}{\small{Classification robustness}}
    & \wrapp{0.2}{\small{Strong class. robustness}}
    \\ 
    \midrule
    \small{Problem domain}
    & \small{\good{General}}
    & \small{\good{General}}
    & \small{\average{Classification}}
    & \small{\average{Classification}}
    \\
    \small{Interpretability}
    & \small{\average{Medium}}
    & \small{\bad{Low}}
    & \small{\good{High}}
    & \small{\average{Medium}}
    \\
    \small{Globally desirable}
    & \small{\yes}
    & \small{\yes}
    & \small{\no}
    & \small{\no}
    \\
    \small{Has loss functions}
    & \small{\yes}
    & \small{\yes}
    & \small{\no}
    & \small{\yes}
    \\
    \midrule
    \small{Adversarial training}
    & \small{\yes}
    & \small{\no}
    & \small{\no}
    & \small{\no}
    \\
    \small{Data augmentation}
    & \small{\no}
    & \small{\no}
    & \small{\yes}
    & \small{\no}
    \\
    \small{Logical-constraint training~\cite{FischerBDGZV19}}
    & \small{\yes}
    & \small{\yes}
    & \small{\no}
    & \small{\yes}
    \\
    \bottomrule
  \end{tabularx}
  \label{tab:comparison}
	\vspace{-1em}
\end{table*}



We finish with a summary of further interesting properties of the four robustness definitions.  
Table~\ref{tab:comparison} shows a summary of all comparison measures considered in the paper.

\textbf{Dataset assumptions}
%
concern the distribution of the training data with respect to the data manifold of the true distribution of inputs, and influence evaluation of robustness.
For SR and LR it is, at minimum, desirable for the network to be robust over the entire data manifold. In the most domains the shape of the manifold is unknown and therefore it is necessary to approximate it by taking the union of the balls around the inputs in the training dataset. We are not particularly interested about whether the network is robust in regions of the input space that lie off the data manifold, but there is no problem if the network is robust in these regions. Therefore these definitions make no assumptions about the distribution of the training dataset.

This is in contrast to CR and SCR. 
Rather than requiring that there is only a small change in the output, they require that there is no change to the classification. This is only a desirable constraint when the region being considered does not contain a decision boundary. Consequently when one is training for some form of classification robustness, one is implicitly making the assumption that the training data points lie away from any decision boundaries within the manifold.
In practice, most datasets for classification problems assign a single label instead of an entire probability distribution to each input point, and so this assumption is usually valid. However, 
for  datasets that contain input points that may lie close to the decision boundaries, CR and SCR may result in a logically inconsistent specification.

%

\textbf{Interpretability.}
One of the key selling points of training with logical constraints is that, by ensuring that the network obeys understandable constraints, it improves the explainability of the neural network. Each of the robustness constraints encode that ``small changes to the input only result in small changes to the output", but the interpretability of each definition is also important.

All of the definitions share the relatively interpretable $\epsilon$ parameter, which measures how large a perturbation from the input is acceptable. Despite the other drawbacks discussed so far, CR is inherently the most interpretable as it has no second parameter. In contrast, SR and SCR require extra parameters, $\delta$ and $\eta$ respectively, which measure the allowable deviation in the output. 
Their addition makes these models less interpretable.

Finally we argue that, although LR is the most desirable constraint, it is also the least interpretable. Its second parameter $L$ measures the allowable change in the output as a proportion of the allowable change in the input. It therefore requires one to not only have an interpretation of distance for both the input and output spaces, but to be able to relate them. In most domains, this relationship simply does not exist. Consider the MNIST dataset, both the commonly used notion of pixel-wise distance used in the input set, although crude, and the distance between the output distributions are both interpretable. However, the relationship between them is not. For example, what does allowing the distance between the output probability distributions being no more than twice the distance between the images actually mean? This therefore highlights a common trade-off between complexity of the constraint and its interpretability.

\section{Conclusions}
\label{sec:discussion}

These case studies have demonstrated the importance of
  emancipating the study of desirable properties of neural networks from a concrete training method, 
  and
  studying these properties in an abstract mathematical way. 
%
  For example,
we have discovered that some  robustness properties can be ordered by logical strength and some are incomparable.
Where ordering is possible, training for a stronger property helps in verifying a weaker property.
Some of the stronger properties, such as Lipschitz robustness, are not yet feasible for the modern DNN solvers, such as Marabou~\cite{marabou}. 
Moreover, we show that the logical strength of the property may not guarantee other desirable properties, such as interpretability.
Some of these findings lead to very concrete recommendations, e.g.: it is best to avoid CR and SCR as they may lead to inconsistencies;
when using LR and SR, one should use stronger property (LR) for training in order to be successful in verifying a weaker one (SR).
In other cases, the distinctions that we make do not give direct prescriptions, but merely discuss the design choices and trade-offs. 

This paper also shows that constraint security, a measure intermediate between constraint accuracy and constraint satisfaction, is a useful tool in the context of tuning the continuous verification loop.
It is more efficient to measure and can show more nuanced trends than constraint satisfaction.
It can be used to tune training parameters and build hypotheses which we ultimately confirm with constraint satisfaction.

We hope that this study will
contribute towards establishing a solid methodology for continuous verification,
by setting up some common principles to unite verification and machine learning approaches to DNN robustness.





%

\bibliographystyle{splncs04}
\bibliography{CAV22}

\clearpage

\appendix

\section{Background Definitions}

\subsection{Neural Networks and their Verification}
A deep neural network (DNN)
is a directed
graph, whose nodes (``neurons'') are organized into layers. The first
layer is called the input layer, and each of its neurons is assigned a
value by the user. Then, values for each of the subsequent, hidden
layers of the network are computed automatically, each time using
values computed for the preceding layer. Finally, values for the
network's final, output layer are computed, and returned to the
user. In classification networks, each output neuron represents a
possible label; and the neuron assigned the highest value represents
the label that the input is classified as.

In recent years, due to the discovery of adversarial perturbations and
other safety and security issues in DNNs, the formal methods
community has been developing methods for verifying their
correctness. In DNN verification, a network $N$ is regarded as a
transformation $N:\mathbb{R}^n\rightarrow\mathbb{R}^m$, and a user
supplies a Boolean precondition $P$ over $\mathbb{R}^n$ and Boolean
postcondition $Q$ over $\mathbb{R}^m$.\footnote{There exist more
  elaborate formulations of the DNN verification problem; we focus
  here on this variant, for simplicity.}
The verification problem is to
determine, given a triple $\langle P, N, Q\rangle$, whether there
exists a concrete input $x_0\in \mathbb{R}^n$ such that $P(x_0)$ and
$Q(N(x_0))$ both hold. The DNN verification problem is NP
complete~\cite{KaBaDiJuKo17Reluplex}, and many approaches have been
devised to solve it in practice
(e.g.~\cite{HuKwWaWu17,SinghGPV19,KaBaDiJuKo17Reluplex}, and many
others). The details of these techniques are mostly beyond our scope
here --- see~\cite{liu2019survey} for a thorough survey.

\section{Experimental Setup: Full Exposition}
\label{sec:appendix}



\textbf{Datasets.}
In our experiments we use two benchmark datasets:
\begin{enumerate}
\item
  The \emph{FASHION MNIST} (or just \emph{FASHION}) dataset~\cite{fashionmnist}  consists of $28 \times 28$ greyscale images of clothing items; 60,000 for training and 10,000 for testing. It has 10 classes: T-shirt/top, Trouser, Pullover, Dress, Coat, Sandal, Shirt, Sneaker, Bag, Ankle boot.
\item
  The \emph{GTSRB} dataset~\cite{gtsrb} contains $50,000$ images of German traffic signs. There are 40 classes and image sizes vary between $15 \times 15$ to $250 \times 250$ pixels. We use a modified version here that contains $12,660$ training images and $4,170$ test images. These images are all centre-cropped, greyscaled, $48\times 48$ pixel and belong to 10 classes.
\end{enumerate}


\subsubsection{Networks Trained to Evaluate Constraint Security}

The networks used in these experiments consist of two fully connected layers: the first one having 100 neurons and ReLU as activation function, and the last one having 10 neurons.
We then apply a clamp function $[-100, 100]$ to the network's output. We don't use the traditional softmax function because the former is compatible with the constraint verification tools such as Marabou whereas the latter is not.
The predicted classification is then taken as the output with the maximum score.

%
For instance, $\mnistfunc = F_0 \circ F_1$, where $F_1 : \real^{784} \rightarrow \real^{100}$ and $F_0 : \real^{100} \rightarrow \real^{10}$,
  $\alpha_0 = ReLU$ and $\alpha_1(x) = \text{clamp}(x, -100, 100)$.

\textbf{Loss functions.}
Since our experiments study classification problems, we will use the cross-entropy loss function as our baseline loss function:
\begin{equation*}
    \label{eq:ce}
    \losssymbol_{ce}(\x, \y) = - \sum_{i=1}^{m} \y_i \; \log(f(\x)_i)
\end{equation*}
For the $ \lossfn_C$ component of the loss function $ \lossfn^*$, we use the constraint-to-loss function translation of~\cite{FischerBDGZV19}. In all experiments, we use the Adam optimiser~\cite{adam_A_method_for_stochastic_optimization_2017_kingma} with the following learning parameters: $\eta = 0.0001$, $100$ epochs, a batch size of $128$.




\begin{wraptable}{R}{0.5\linewidth}
  \caption{\emph{\footnotesize{Standard test set accuracy (as \% of the dataset instances) for chosen trained networks.}}}
  \begin{tabularx}{\linewidth}{p{3.7cm} p{1.2cm} p{1.2cm}}
    \toprule
    \small{Training Regime:}
    & \small{FASHION}
    & \small{GTSRB}
    \\
    \midrule
   \small{Baseline}
    &  \small{88.2}
    &   \small{92.4}
    \\
    \small{Data Augmentation (RU)}
    &  \small{88.6}
    &  \small{92.8}
    \\
     \small{Data Augmentation (FGSM)}
    &  \small{88.8}
    &  \small{94.5}
    \\
     \small{Adversarial Training}
    &  \small{85.1}
    &  \small{83.5}
    \\
     \small{Constraint Loss (SR)}
    &  \small{88.2}
    &  \small{93.3}
    \\
     \small{Constraint Loss (SCR)}
    &  \small{88.1}
    &  \small{91.9}
    \\
     \small{Constraint Loss (LR)}
    &  \small{86.6}
   &  \small{93.1}
    \\
    \bottomrule
  \end{tabularx}
  \label{tab:accuracy}
	\vspace{-2em}
\end{wraptable}

\textbf{Settings.}
We keep the architecture the same throughout the experiments, and only vary the training. Thus our \emph{Baseline} network is trained just with cross-entropy.
Data Augmentation adds 2 additional images in the $0.1$-$\epsilon$-ball around each image, and it samples either randomly from a uniform distribution (RU) or using an FGSM attack.
Adversarial training refers the training procedure described in Section~\ref{sec:robustness-training}, with PGD sampling.
In all other cases, we use a constraint loss function $ \lossfn_C$ defined as in~\cite{FischerBDGZV19}, and we use the constraints SR, SCR, LR as defined in Section~\ref{sec:robustness-training} with $\alpha = 1$, $\beta = 0.2$, $\epsilon = 0.1$, $\delta = 10$, $\eta = 0.52$, $L = 10$ and $L^{\infty}$ distance metrics. All networks trained with  $ \lossfn^*$ use sampling by the PGD attack, for efficiency (as well as comparability).

We start with noting the standard test set accuracy of the resulting neural networks in Table~\ref{tab:accuracy}, making sure that our different training regimes do not deteriorate networks' general performance too drastically. The most notable accuracy drop occurs for adversarial training.  

\subsubsection{Networks Trained to Evaluate Constraint Satisfaction in Marabou}


The networks used in these experiments are the same as before, except for the first layer that has 30 neurons instead of 100. This shrinking was to facilitate the use of the Marabou framework.
We then consider, for each of the two datasets, four networks:
\begin{enumerate}
\item The \emph{baseline} network, trained just with cross-entropy.
\item The \emph{SCR} network, trained using the \emph{SCR} constraint with $\alpha = 1$, $\beta = 0.2$, $\epsilon = 0.1$, $\eta = 0.52$ and $L^{\infty}$.
\item The \emph{SR} network, trained using the \emph{SR} constraint with $\alpha = 1$, $\beta = 0.2$, $\epsilon = 0.1$, $\delta = 10$ and $L^{\infty}$.
\item The \emph{LR} network, trained using the \emph{LR} constraint with $\alpha = 1$, $\beta = 0.2$, $\epsilon = 0.1$, $L = 10$ and $L^{\infty}$.
\end{enumerate}

Finally, we define three queries for evaluating constraint satisfaction: Classification, Standard and Lipschitz Constraints as defined in Section~\ref{sec:robustness-training} with $\alpha = 1$, $\beta = 0.2$, $\epsilon = 0.1$, $\delta = 10$, $L = 10$ and $L^{\infty}$ distance metrics.
The results of these experiments are reported in Table~\ref{tab:marabou_results} and are calculated over 200 images from the test sets.

\section{Complete Results}
\label{sec:appendixB}

The rest of this appendix will present all the results that, due to space, were not possible to include in the main text. We will show in detail how we systematically evaluated all the trends that we reported. Figures~\ref{fig:SR},~\ref{fig:SR2} are the same as Figures~\ref{fig:modes},~\ref{fig:CL}, and Figures~\ref{fig:SR3}A,~\ref{fig:LR3}A correspond to Figure~\ref{fig:SL-LR}, we report them again here for completeness.


\subsection{Data augmentation vs adversarial training vs training with constraint loss}

\begin{figure*}[ht]
	\centering
	\begin{subfigure}{.8\linewidth}
		\includegraphics[width=1\linewidth]{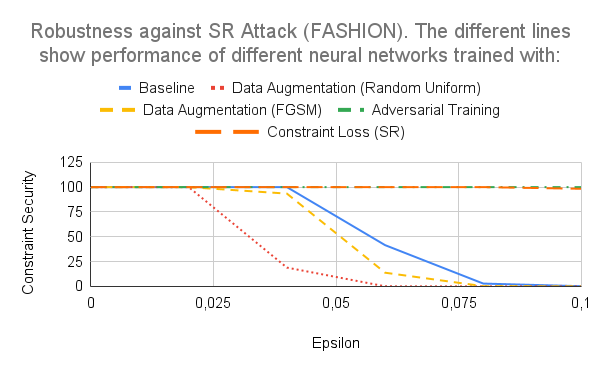}
	\end{subfigure}
	\hspace{0.03\textwidth}
	\begin{subfigure}{.8\linewidth}
		\includegraphics[width=1\linewidth]{figures/GTSRB_AR_.png}
	\end{subfigure}
	\vspace{-1em}
	\caption{\emph{\footnotesize{Experiments that show how adversarial training, training with data augmentation, and training with constraint loss affect standard robustness of neural networks, for varying sizes of the PGD attack (measured by $\epsilon$ values).}}}
	\label{fig:SR}
\end{figure*}

\begin{figure*}[ht]
	\centering
	\begin{subfigure}{.8\linewidth}
		\includegraphics[width=1\linewidth]{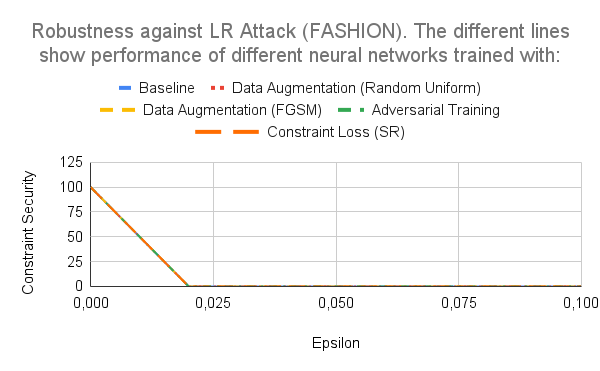}
	\end{subfigure}
	\hspace{0.03\textwidth}
	\begin{subfigure}{.8\linewidth}
		\includegraphics[width=1\linewidth]{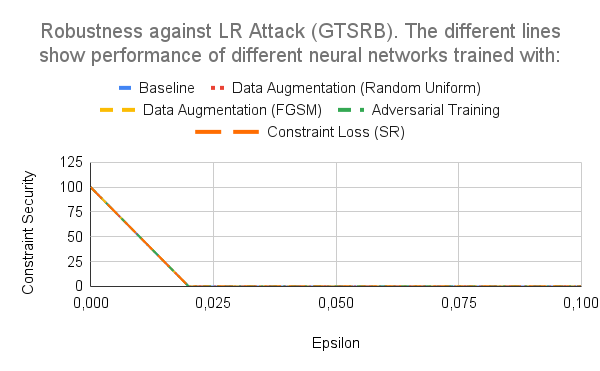}
	\end{subfigure}
	\vspace{-1em}
	\caption{\emph{\footnotesize{Experiments that show how adversarial training, training with data augmentation, and training with constraint loss affect lipschitz robustness of neural networks, for varying sizes of the PGD attack (measured by $\epsilon$ values).}}}
	\label{fig:LR}
\end{figure*}

\begin{figure*}[ht]
	\centering
	\begin{subfigure}{.8\linewidth}
		\includegraphics[width=1\linewidth]{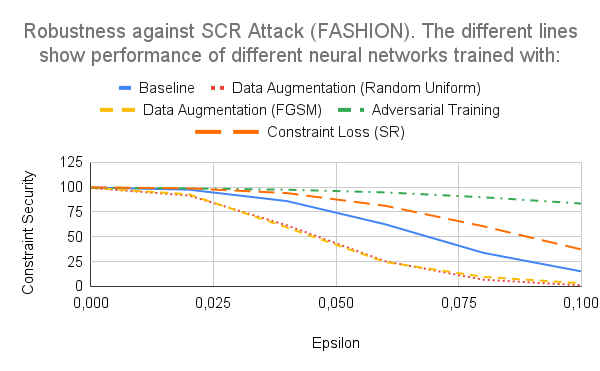}
	\end{subfigure}
	\hspace{0.03\textwidth}
	\begin{subfigure}{.8\linewidth}
		\includegraphics[width=1\linewidth]{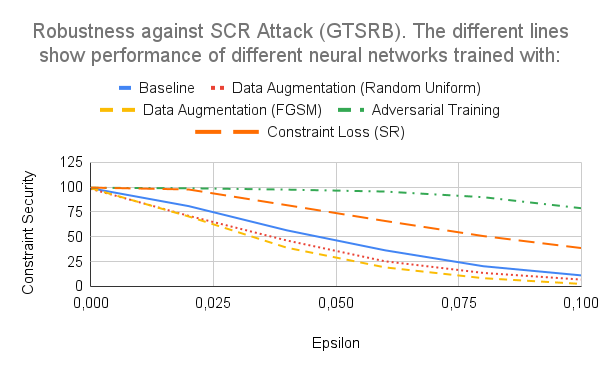}
	\end{subfigure}
	\vspace{-1em}
	\caption{\emph{\footnotesize{Experiments that show how adversarial training, training with data augmentation, and training with constraint loss affect strong classification robustness of neural networks, for varying sizes of the PGD attack (measured by $\epsilon$ values).}}}
	\label{fig:ACR}
\end{figure*}

Figures~\ref{fig:SR},~\ref{fig:LR} and~\ref{fig:ACR} show how the networks trained with the different methods are robust against attacks.
For SR and SCR attacks, Adversarial Training improves significantly the robustness of the model while training with Constraint Loss (SR) also improves it, although not as much.
Training with both the Data Augmentation methods actually reduce robustness of the network.
None of the training techniques succeed in ensuring robustness against LR attacks, indicating that as discussed in Section 3.1, LR is a strong property to hold.
Nonetheless we report the graphs for completeness.

\subsection{Training with different constraint losses}

\begin{figure*}[ht]
	\centering
	\begin{subfigure}{.8\linewidth}
		\includegraphics[width=1\linewidth]{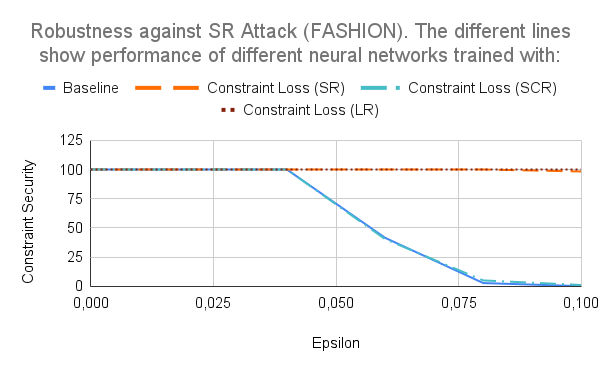}
	\end{subfigure}
	\hspace{0.03\textwidth}
	\begin{subfigure}{.8\linewidth}
		\includegraphics[width=1\linewidth]{figures/GTSRB_SR_.png}
	\end{subfigure}
	\vspace{-1em}
	\caption{\emph{\footnotesize{Experiments that show how different choices of a constraint loss affect standard robustness of neural networks, for varying sizes of the PGD attack (measured by $\epsilon$ values).}}}
	\label{fig:SR2}
\end{figure*}

\begin{figure*}[ht]
	\centering
	\begin{subfigure}{.8\linewidth}
		\includegraphics[width=1\linewidth]{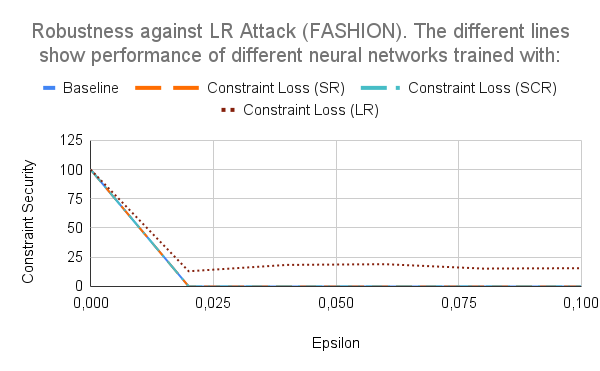}
	\end{subfigure}
	\hspace{0.03\textwidth}
	\begin{subfigure}{.8\linewidth}
		\includegraphics[width=1\linewidth]{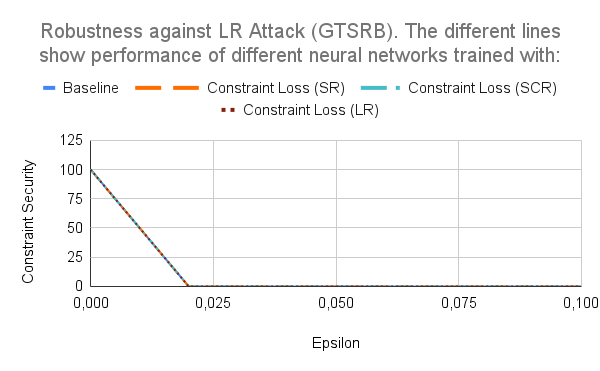}
	\end{subfigure}
	\vspace{-1em}
	\caption{\emph{\footnotesize{Experiments that show how different choices of a constraint loss affect lipschitz robustness of neural networks, for varying sizes of the PGD attack (measured by $\epsilon$ values).}}}
	\label{fig:LR2}
\end{figure*}

\begin{figure*}[ht]
	\centering
	\begin{subfigure}{.8\linewidth}
		\includegraphics[width=1\linewidth]{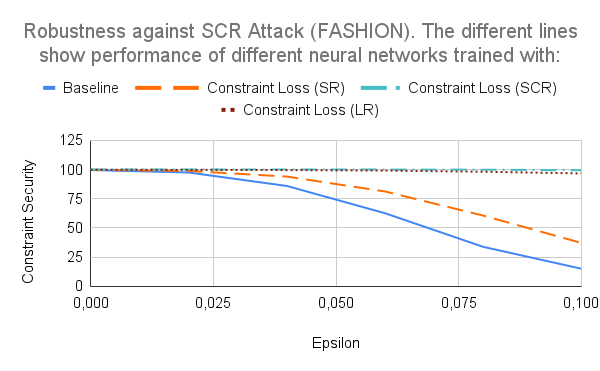}
	\end{subfigure}
	\hspace{0.03\textwidth}
	\begin{subfigure}{.8\linewidth}
		\includegraphics[width=1\linewidth]{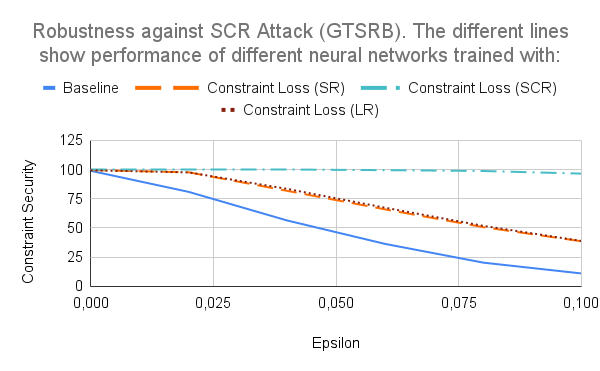}
	\end{subfigure}
	\vspace{-1em}
	\caption{\emph{\footnotesize{Experiments that show how different choices of a constraint loss affect strong classification robustness of neural networks, for varying sizes of the PGD attack (measured by $\epsilon$ values).}}}
	\label{fig:ACR2}
\end{figure*}

Figures~\ref{fig:SR2},~\ref{fig:LR2} and~\ref{fig:ACR2} show how the networks trained with the different constraint losses are robust against attacks.
For SR and SCR attacks, we can see a trend in the networks trained with the same constraint that are generally more robust against the respective attacks.
On the other hand, models trained with LR are generally more robust and they have the best average improvement against all attacks.
None of the training techniques succeed in ensuring robustness against LR attacks, except the network trained with LR on the FASHION dataset.

\subsection{Networks' behaviours against different attacks}

\begin{figure*}[ht]
	\centering
	\begin{subfigure}{.8\linewidth}
		\includegraphics[width=1\linewidth]{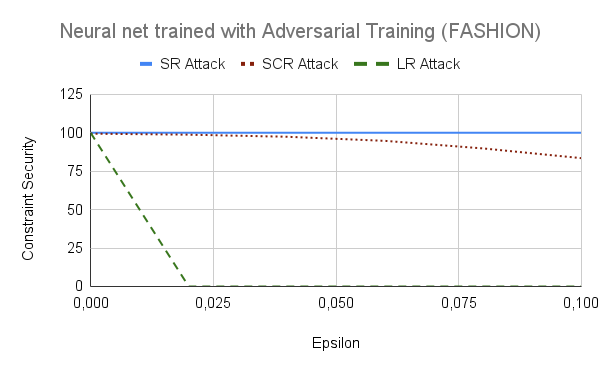}
	\end{subfigure}
	\hspace{0.03\textwidth}
	\begin{subfigure}{.8\linewidth}
		\includegraphics[width=1\linewidth]{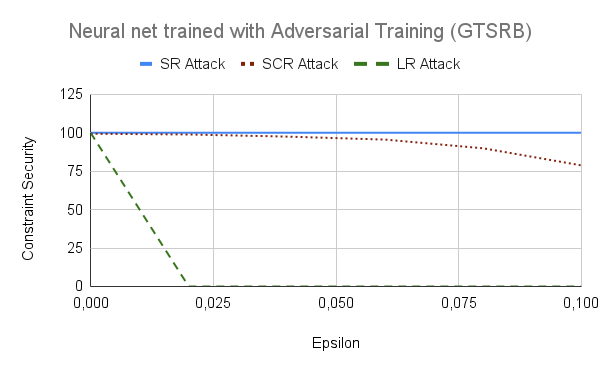}
	\end{subfigure}
	\vspace{-1em}
	\caption{\emph{\footnotesize{Experiments that show how the networks trained with Adversarial Training perform when evaluated against different definitions of robustness underlying the attack.}}}
	\label{fig:AT}
\end{figure*}

\begin{figure*}[ht]
	\centering
	\begin{subfigure}{.8\linewidth}
		\includegraphics[width=1\linewidth]{figures/FASHION_SR_3.png}
	\end{subfigure}
	\hspace{0.03\textwidth}
	\begin{subfigure}{.8\linewidth}
		\includegraphics[width=1\linewidth]{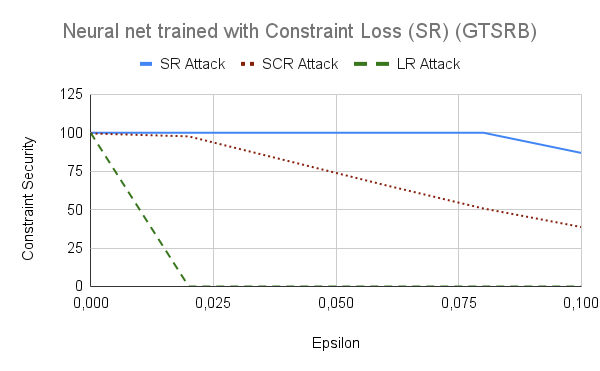}
	\end{subfigure}
	\vspace{-1em}
	\caption{\emph{\footnotesize{Experiments that show how the networks trained with SR constraints perform when evaluated against different definitions of robustness underlying the attack.}}}
	\label{fig:SR3}
\end{figure*}

\begin{figure*}[ht]
	\centering
	\begin{subfigure}{.8\linewidth}
		\includegraphics[width=1\linewidth]{figures/FASHION_LR_3.png}
	\end{subfigure}
	\hspace{0.03\textwidth}
	\begin{subfigure}{.8\linewidth}
		\includegraphics[width=1\linewidth]{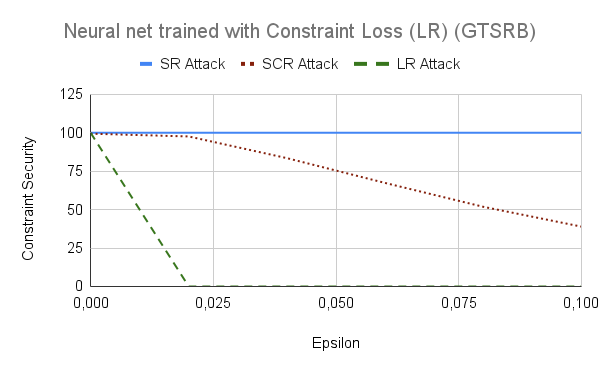}
	\end{subfigure}
	\vspace{-1em}
	\caption{\emph{\footnotesize{Experiments that show how the networks trained with LR constraints perform when evaluated against different definitions of robustness underlying the attack.}}}
	\label{fig:LR3}
\end{figure*}

Previous experiments show that the most robust models are the ones trained with Adversarial Training, LR and SR.
Figures~\ref{fig:AT},~\ref{fig:SR3} and~\ref{fig:LR3} select the relevant data from previous experiments to provide a comparison of how these models perform against the different attacks.
All the networks struggle the most against LR attacks, while they present significant robustness against SR attacks and a slightly less but still important robustness against SCR attacks.
Overall we can see that Adversarial Training provides the best defence against SR and SCR attacks, immediately followed by training with Constraint Loss LR. However, as already reported above, the only network that show some robustness against LR attacks is the one trained with Constraint Loss LR.

\end{document}